\documentclass[conference]{IEEEtran}
\IEEEoverridecommandlockouts
\usepackage{cite}
\usepackage{amsmath,amssymb,amsfonts}
\usepackage{algorithmic}
\usepackage{graphicx}
\usepackage{subfigure}
\usepackage{textcomp}
\usepackage{xcolor}
\def\BibTeX{{\rm B\kern-.05em{\sc i\kern-.025em b}\kern-.08em
    T\kern-.1667em\lower.7ex\hbox{E}\kern-.125emX}}
\begin{document}

\title{X-MLP: A Patch Embedding-Free MLP Architecture for Vision}

\author{\IEEEauthorblockN{Xinyue Wang, Zhicheng Cai and Chenglei Peng}
\IEEEauthorblockA{\textit{School of Electronic Science and Engineering, Nanjing University}
\\ Email: \{201180089, caizc\}@smail.nju.edu.cn, pcl@nju.edu.cn}
}

\maketitle

\begin{abstract}
Convolutional neural networks (CNNs) and vision transformers (ViT) have obtained great achievements in computer vision. Recently, the research of multi-layer perceptron (MLP)  architectures for vision have been popular again. Vision MLPs are designed to be independent from convolutions and self-attention operations. However, existing vision MLP architectures always depend on convolution for patch embedding. Thus we propose X-MLP, an architecture constructed absolutely upon fully connected layers and free from patch embedding. It decouples the features extremely and utilizes MLPs to interact the information across the dimension of width, height and channel independently and alternately. X-MLP is tested on ten benchmark datasets, all obtaining better performance than other vision MLP models. It even surpasses CNNs by a clear margin on various dataset. Furthermore, through mathematically restoring the spatial weights, we visualize the information communication between any couples of pixels in the feature map and observe the phenomenon of capturing long-range dependency.
\end{abstract}

\section{Introduction}
As the first end-to-end vision model, convolutional neural networks  (CNNs)~\cite{krizhevsky2012imagenet,simonyan2014very,he2016deep} have been the de-facto standard tool in computer vision field~\cite{chen2021you,deng2021fractal,bai2020adaptive} for a long duration. Taking advantage of the self-attention which originally raised in the natural language processing~\cite{vaswani2017attention} , vision transformers (ViT) ~\cite{dosovitskiy2020image,touvron2021going} attained marvelous performance and surpassed CNNs in many vision tasks ~\cite{liu2021swin,carion2020end,wang2021end}. Recently, multi-layer perceptron (MLP) architectures have been proved to achieve comparable performance to CNNs and ViTs ~\cite{melas2021you,tolstikhin2021mlp,touvron2021resmlp} under certain circumstances.
Regarded as a competitive but conceptually and technically simple alternative to convolution and self-attention, vision MLPs inherit the trend that abandons the paradigm of hand-designed visual features and inductive biases and continues the end-to-end learning pattern~\cite{tolstikhin2021mlp}. Thus, the closed loop form MLP to CNN, then to self-attention, and finally back to MLP has been established.

Vision MLPs are designed to be based on pure fully connected layers and similar to ViT architectures which take split patches of images as input. However, it is conventionally for existing vision MLPs like MLPMixer~\cite{tolstikhin2021mlp} and ResMLP~\cite{touvron2021resmlp} to depend on convolution operation to conduct the patch embedding. This process runs counter to the original purpose of constructing a vision architecture with pure fully connected layers. Thus we propose the X-MLP, which means extreme MLP for vision. Unlike MLPMixer or other existing vision MLPS which utilize convolutional patch embedding to product per-location 2-D feature maps shaped as ``patches $\times$ channels''  and adopt MLPs repeatedly across spatial locations and feature channels, our X-MLP entirely utilizes linear projections to interact the information across the dimension of width, height and channels in the 3-D feature maps individually and alternately. Consequently, X-MLP is a MLP architecture free from convolutional patch embedding and stacked by fully connected layers purely and entirely.
 
\begin{figure*}[tbp]
   \centering
   \includegraphics[width=0.95\textwidth]{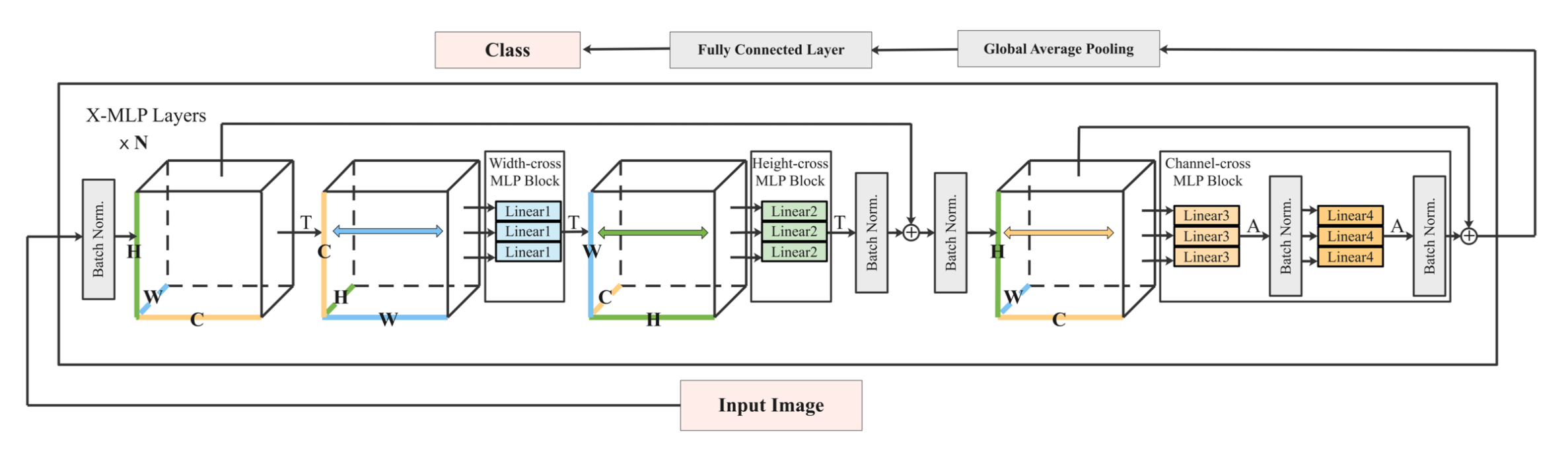}

   \caption{Basic macro-structure of X-MLP. X-MLP takes the original image as the input, if consists of X-MLP layers and a classifier. The X-MLP layers contains three MLP blocks: \emph{width-cross} MLPs, \emph{height-cross} MLPs, and \emph{channel-cross} MLPs, interacting the information across the dimension of width, height, and channel respectively. In the figure, \textbf{T} stands for transpose and \textbf{A} stands for activation, which is PReLU in the model}
   \label{p1}
\end{figure*}
 
Fig.~\ref{p1} exhibits the basic macro-structure of X-MLP. X-MLP takes the original images as the input. Because of the linear projections conducted respectively across three dimensions, the feature maps produced in each layer possess a shape of ``width $\times$ height $\times$ depth (also referred to as \emph{channels})''. X-MLP is specifically constructed by X-MLP layers. One X-MLP layer employs three types of MLP layers in order, that is, \emph{width-cross} MLPs, \emph{height-cross} MLPs, and \emph{channel-cross} MLPs. 
The \emph{width-cross} MLPs interact the information across the dimension of width. They operate on each vector parallel to the dimension of width in the input feature maps independently and take individual vectors as inputs. Similarly, the \emph{height-cross} MLPs allow information communication across the dimension of height. They operate on each vector perpendicular to the plane of `` width $\times$ depth'' in the input feature maps independently and take individual vectors as inputs. Combining these two MLPs successively is supposed to extract the spatial features in certain channel.  
The \emph{channel-cross} MLPs operate on each position in the feature maps independently and take one individual pixel in the input feature maps once as the input. This MLP layer communicates the information across the channel dimension in certain spatial position. The learned parameters of these MLPs are all shared by each input vector. Moreover, The numbers of hidden neurons are not constant in different layers. Actually, the \emph{width-cross} MLPs and the \emph{height-cross} MLPs can be utilized to customizedly adjust the spatial size of the feature maps, and the \emph{channel-cross} MLPs can adjust the depth of the feature maps. 
Our X-MLP also introduces the residual connections and batch normalization into the micro-architecture of X-MLP layers. Finally, the feature maps are average pooled globally across the channel dimension, and input into a linear classifier.   

Compared to other existing vision MLPs which bear resemblance to the vision transformers, our X-MLP is more likely to the CNNs, more specifically, the \emph{depth-wise separable} convolution. Like the \emph{depth-wise} convolution, the \emph{width-cross} MLPs and the \emph{height-cross} MLPs are combined to extract the spatial information in certain channels. However, these two MLPs are supposed to capture long-range dependencies instead of capturing local features. In terms of the \emph{channel-cross} MLPs, they are actually the same as the \emph{point-wise} convolution which aggregates features across different channels in certain location. In addition, the shape alter of the feature maps produced in each X-MLP layer follows the design in CNN models, instead of being constant like other vision MLPs do.  

In summary, the main contributions of this paper are listed here:
\begin{itemize}
    \item X-MLP is the first vision MLP that casts off convolutional patch embedding.  Thus the model is totally free from convolution and self-attention. This patch embedding-free design enlarges the family of MLPs for vision.
    \item X-MLP is constructed upon fully connected layers entirely. It bears resemblance to CNN models, but it is more simple and capable of capturing long-range dependencies.
    \item X-MLP decouples the features extremely. It extracts and mixes the features from three dimensions respectively.
    \item To explore the pure effectiveness of different models, we tested X-MLP and other compared models on ten benchmark datasets without pre-training or heavy data augmentations. X-MLP attained better performance than other MLPs on all the datasets with fewer parameters. X-MLP even surpassed CNN models significantly on most cases.
    \item Because of the design that utilizing linear layers to interact information across three dimensions independently, by combining the \emph{width-cross} MLPs and the \emph{height-cross} MLPs, we can make observations on the spatial information communication between any couples of pixels in the inputs and the phenomenon of capturing long-range dependency. Besides, we can observe the relationship between any couples of elements in the feature maps by additionally taking \emph{channel-cross} MLPs into consideration.
\end{itemize}

\section{Model Architecture of X-MLP}

The key point of modern vision models lies in how to extract and aggregate both spatial and channel features. It can be classified into two types. One of them extract spatial and channel information at once, containing CNNs whose the size of the standard convolutional kernels is larger than $1$, ViT, and other self-attention based models ~\cite{ramachandran2019stand,zhao2020exploring}. The other extract spatial and channel information separably, containing \emph{depth-wise separable} convolution~\cite{howard2017mobilenets} which combines \emph{depth-wise} convolution and \emph{point-wise} convolution together and former vision MLPs which separates the per-location and cross-location operations. However, X-MLP conducts the separation further. It splits spatial features into features of width and height dimensions and operates communication across these two dimensions respectively and alternately. As a result, the spatial features are supposed to be extracted and aggregated globally.

\subsection{Prototype of X-MLP}
In this section, we will detail the prototype of the X-MLP. The overall structure of the basic model is exhibited in Fig.~\ref{p1}. X-MLP takes the original image as the input. This approach makes the model free from splitting the image into patches and accordingly linear projections. X-MLP is stacked on X-MLP layers. The feature maps produced in each X-MLP layer are three-dimensional real-valued tensors. Fig.~\ref{p2-a} illustrates the architecture of the basic X-MLP layers more concisely and specifically. At the end of the model, there is a global average pooling layer and a linear layer for classification. 

\textbf{Basic X-MLP Layer.} Suppose $\textbf{X} \in \mathbb{R}^{C\times H\times W}$ is the input feature map and $\textbf{Y} \in \mathbb{R}^{C^{\prime}\times H^{\prime}\times W^{\prime}}$ is the feature map output by X-MLP layer. Each basic X-MLP layer mainly consists of three MLP blocks. The first one is the  \emph{width-cross} MLP block, which operates on vectors $\textbf{X}_{c,h,:}$ for each position $(c,h) \in \mathbb{R}^{C\times H}$ and maps $\mathbb{R^{W}}\mapsto \mathbb{R^{W^{\prime}}}$. It intends to interact the information across the dimension of width. The following one is the  \emph{height-cross} MLP block, which operates on vectors $\textbf{X}_{c,:,w}$ for each position $(c,w) \in \mathbb{R}^{C\times W}$ and maps $\mathbb{R^{H}}\mapsto \mathbb{R^{H^{\prime}}}$. It allows the information communication across the dimension of height. These two MLP blocks are complementary and combined to restore and extract the spatial features globally. The final one is the  \emph{channel-cross} MLP block, operating on vectors $\textbf{X}_{:,h,w}$ for each pixel $(h,w) \in \mathbb{R}^{H\times W}$, mapping $\mathbb{R^{C}}\mapsto \mathbb{R^{C^{\prime}}}$. It is utilized to aggregate the features from different channels in certain pixel. All the learned parameters in these three MLP blocks are shared across these operated vectors.  In the basic X-MLP layers, the \emph{width-cross} and \emph{height-cross} MLP blocks contain one fully connected layer without any activation or normalization layers, while \emph{channel-cross} MLP blocks contain two fully connected layers followed by one activation layer and one batch normalization layer  as a common practice~\cite{bjorck2018understanding}. The configurations are different in improved X-MLP layers which will be stated in Section~\ref{s1}. The basic X-MLP layer can be written mathematically as Eq.~\ref{eq1}:

\begin{equation}
    \begin{split}
    & \textbf{U}_{c,h,:} = \textbf{W}_1BN(\textbf{X})_{c,h,:} \\
    & \textbf{V}_{c,:,w} = \textbf{W}_2\textbf{U}_{c,:,w} \\
    & \textbf{O}_{c,h,w}= BN(BN(\textbf{X}) + BN(\textbf{V}))_{c,h,w} \\
    & \textbf{Y}_{:,h,w} = BN(\sigma(\textbf{W}_4BN(\sigma(\textbf{W}_3\textbf{O}_{:,h,w})))) + \textbf{O}_{:,h,w} \\
    &\textbf{where}\ \ c \in [1,C],\ \ h \in [1,H],\ \ w \in [1,W]
    \label{eq1}
    \end{split}
\end{equation}

Where the $\sigma$ stands for the non-linear activation function, which is PReLU ~\cite{he2015delving} utilized in X-MLP. $BN$ stands for the batch normalization layer, which is validated to be more beneficial than other normalization layers empirically. Here, $\textbf{W}_1 \in \mathbb{R}^{W\times W^{\prime}}$, $\textbf{W}_2 \in \mathbb{R}^{H\times H^{\prime}}$, $\textbf{W}_3 \in \mathbb{R}^{C\times \textbf{$\epsilon$} C^{\prime}}$, and $\textbf{W}_4 \in \mathbb{R}^{\textbf{$\epsilon$} C^{\prime}\times C^{\prime}}$ stand for the learned weights in the four fully connected layers respectively, and \textbf{$\epsilon$} stand for the expansion factor, which is set to be $4$ constantly in our paper. In addition to the MLP layers and batch normalization layers, residual connections are adopted in the X-MLP model when the size of the input feature map is equivalent to that of the output feature map.

\textbf{Pyramidal Structure.} Contrary to that the sizes of the feature maps are set to be a constant in existing vision MLPs, they can be adjusted by these fully connected layers in X-MLP layers. As a matter of fact, the sizes of the feature maps in X-MLP have a pyramidal structure, which is the same as the typical design of CNNs ~\cite{simonyan2014very,he2016deep,han2017deep}. Namely, the deeper layers possess more feature channels and a lower resolution input, which has been proved to be more beneficial to the actual learning progress. Moreover, because of the decrease of the spatial size and the gradual increase of the feature channels, X-MLP models have less learned parameters than existing vision MLPs. 

\textbf{Decouple the Features Extremely.} As mentioned above, X-MLP attempts to extract the features in an extreme way. Existing vision models extract spatial and channel features either simultaneously or separately. However, on the basis of extracting spatial and channel features respectively, X-MLP further decouples the spatial features into the features across the width and height dimensions.  Through interacting the complementary information across these two dimensions by employing \emph{width-cross} and \emph{height-cross} MLP blocks in order repeatedly, it is supposed to restore and aggregate spatial features. Besides, because of the pyramidal structure of the feature maps, the number of the learned parameters in these two MLP blocks are far less than that in the \emph{channel-cross} MLP blocks. 

\textbf{Capture Global Long-range Dependencies.} In addition, taking advantages of these MLP blocks enables the model to capture the global long-range dependencies. To state the reason, given two pixel $X_{i,j}$ and $ X_{m,n}$, where $i\neq m$ and $j\neq n$ (omitting the channel dimension). First, we interact the information across the width. Consequently, $X_{i,j}$ is attached to the pixel $X_{i,n}$. Then, the information across the height is communicated. Thus $X_{i,n}$ is attached to the pixel $X_{m,n}$. As a result, the 
connection between pixel $X_{i,j}$ and $ X_{m,n}$ has been established by pixel $X_{i,n}$. Pixel $X_{m,j}$ also makes the contribution accordingly. This connection will be strengthened by operating alternately as the layer of network deepens. Moreover, MLP blocks make the model sensitive to the spatial locations in the input images.

\textbf{Amount of Parameters.} To compared the number of learned parameters with the standard convolutional layer, suppose that the input feature map $\textbf{X} \in \$\mathbb{R}^{H\times W\times C}$ and the kernel size is set to be $K$ ($K>3$). The parameter number of convolution is $N_C=K\times K\times C\times C = K^2C^2$. For X-MLP, it is $N_X=H^2+ W^2+C^2$. Suppose that $H=W$, that is, $N_X=2H^2+C^2$. It is obviously that when $\frac{H}{C}\!<\!\sqrt{\frac{K^2-1}{2}}$, the parameter number of X-MLP layer is smaller than that of convolutional layer. For feature maps with a pyramidal structure, it always meets the condition.

In terms of extracting the spatial features, compared with ordinary fully connected layers , the number of learned parameters in one fully connected layer is $H\times W\times H\times W = H^4$. For X-MLP, it is $2H^2$. As a common practice, $H$ is larger than 1. So the number of weights in X-MLP is $\frac{1}{2}H^2$ times less than that in fully connected layer.

\textbf{Analysis of Computational Complexity.}To compared the computational complexity with the standard convolutional layer, still we suppose that the input feature map $\textbf{X} \in \$\mathbb{R}^{H\times W\times C}$ and the kernel size is set to be $K$ ($K>3$). The complex of convolution operation is $\mathcal{O}(H\times W\times C\times C\times K\times K)=\mathcal{O}(H^2C^2K^2)$. For X-MLP, it is $\mathcal{O}(H\times W\times C\times H +H\times W\times C\times W +H\times W\times C\times C )=\mathcal{O}(2H^2C+H^2C^2)$, which the term of $\mathcal{O}(2H^2C)$ is negligible compared to  $\mathcal{O}(H^2C^2)$. Thus the computational complexity of X-MLP is $\frac{1}{K^2}$ of convolution operation.

In terms of extracting the spatial features, compared with ordinary fully connected layers , the computational complexity of one fully connected layer is $\mathcal{O}(H^4C)$. For X-MLP, it is $\mathcal{O}(2H^2C)$. As a common practice, $H$ is larger than 1. So the computational complexity of X-MLP is $\frac{H^2}{2}$ times less than fully connected layer.

\subsection{Architectures of Improved X-MLP Layers} \label{s1}
To further enhance the performance of X-MLP, we explore some improved architectures based on the basic X-MLP layer.

\begin{figure*}[!ht]
   \centering
   \subfigure[X-Basic Layer]{
      \includegraphics[width=0.95\textwidth]{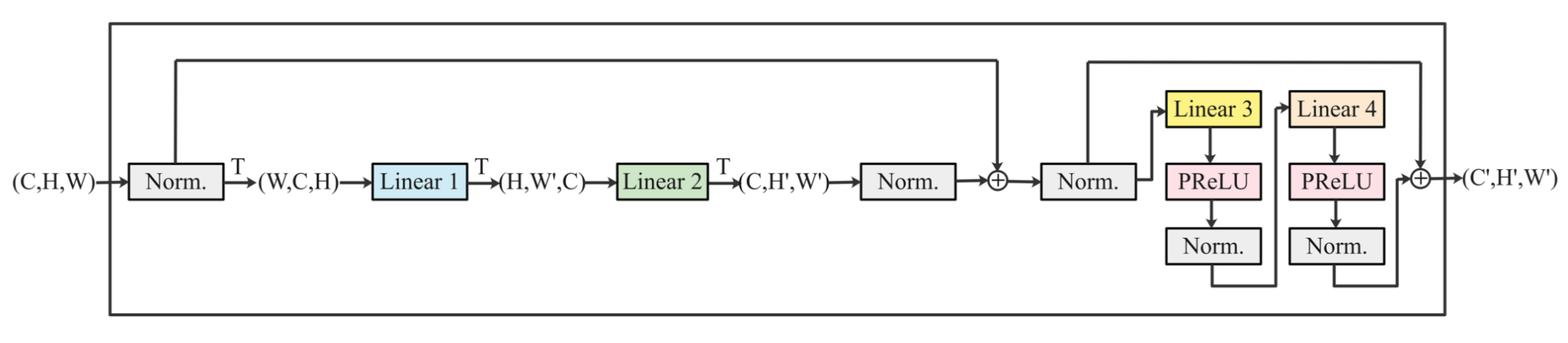}
      \label{p2-a}
      }
   \subfigure[X-Expansion Layer]{
      \includegraphics[width=0.95\textwidth]{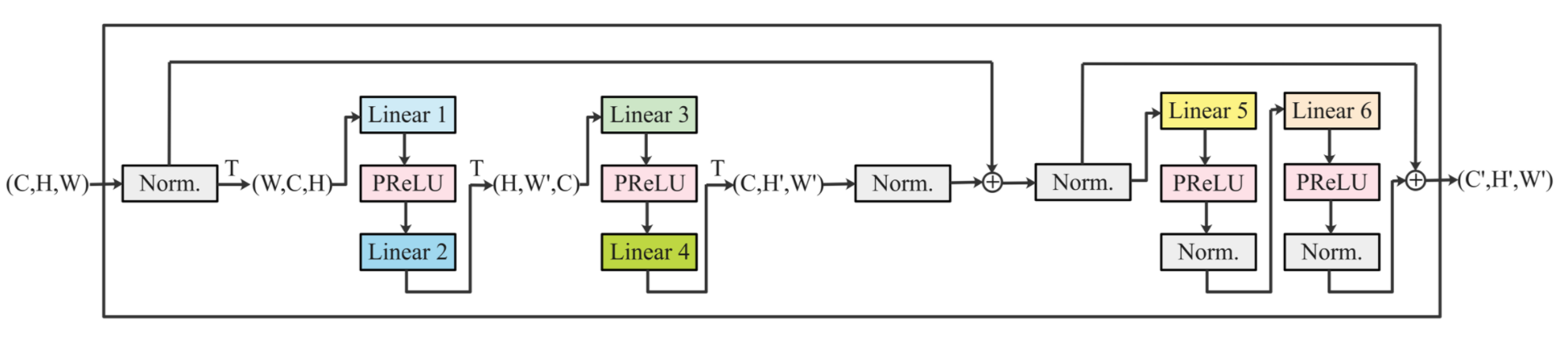}
      \label{p2-b}
      }
   \subfigure[X-Alternate Layer]{
      \includegraphics[width=0.95\textwidth]{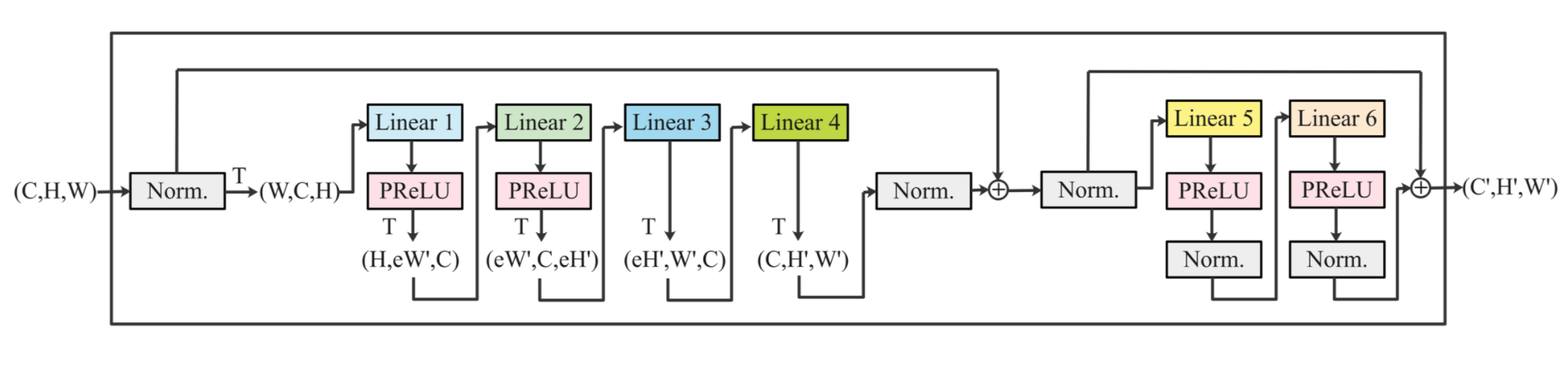}
      \label{p2-c}
      }
   \subfigure[X-Superior Layer]{
      \includegraphics[width=0.95\textwidth]{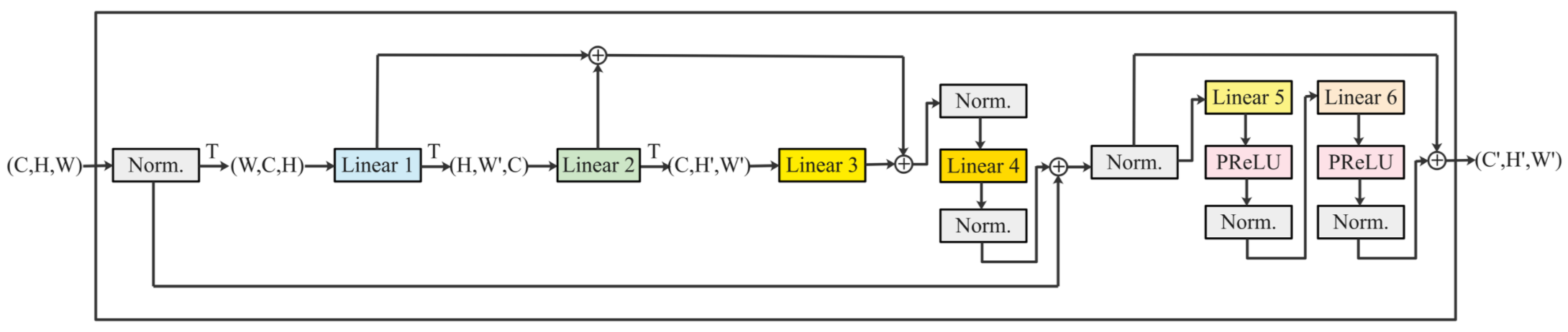}
      \label{p2-d}
      }
   \caption{Architectures of X-MLP layers}
   \label{p2}
\end{figure*}

\textbf{X-Expansion. }In this architecture, we add one more fully connected layer in both \emph{width-cross} and \emph{height-cross} MLP blocks. Besides, we add one non-linear activation function between the two fully connected layers in these MLP blocks. The \emph{channel-cross} MLP blocks are kept unchanged here. Fig.~\ref{p2-b} illustrates the architecture of the X-Expansion layers. This design is supposed to capture the non-linear relationships and enhance ability of extracting spatial features. Compared to the \emph{channel-cross} MLP blocks, these MLP blocks have expansion factor for the hidden layer but are not equipped with batch normalization layers. 



\begin{table*}[!ht]
  \caption{Description of the ten benchmark datasets}
   \label{t1}
  \centering
   \begin{tabular}{|l|c|c|c|c|c|c|c|}
   \hline
   Dataset	& Classes  &	Num. Train& Num. Test &	\ \ \ \ Orig. Size \ \ \ \ &	\ \ \ \ Input size\ \ \ \  & Epochs & Data Aug. \\
   \hline
   Tiny ImageNet$^*$ & 200 &	100,000& 10,000&	$ 64\times64\times3$ &	$64\times64\times3$ & 120 & Rotate, Flip \\
   \hline
   Caltech-256$^*$ & 257&	21,425&	9,182&	various&	$64\times64\times3$ & 80& Rotate, Flip \\
   \hline
   Flowers-102$^*$ &102	&2,040&	6,149&	various	&$64\times64\times3$ & 80& Rotate, Flip \\
   \hline
   Food-101$^*$ & 101&	75,750&	25,250&	various	&$64\times64\times3$  & 80& Rotate, Flip \\
   \hline
   CIFAR-100 & 100&	50,000&	10,000	&$32\times32\times3$&	$64\times64\times3$ & 120& Crop, Flip \\
   \hline
   CIFAR-10 & 10&	50,000&	10,000&	$32\times32\times3$&	$32\times32\times3$& 120& Crop, Flip \\
   \hline
   SVHN & 10&73,257	&26,032	&$32\times32\times3$&	$32\times32\times3$ & 60 & --- --- \\
   \hline
   MNIST & 10&	60,000&	10,000&	$28\times28\times1$&	$32\times32\times1$& 40  & --- --- \\
   \hline
   KMNIST & 10&	60,000&	10,000	&$28\times28\times1$&	$32\times32\times1$& 60 & --- --- \\
   \hline
   Fashion-MNIST &10&	60,000&	10,000&	$28\times28\times1$&	$32\times32\times1$ & 60 & --- --- \\
   \hline
   \end{tabular}
\end{table*}
\textbf{X-Alternate.} In this architecture, we tear the two fully connected layers respectively in the \emph{width-cross} and \emph{height-cross} MLP blocks of X-Expansion layer apart and alternate these four layers in order. The \emph{channel-cross} MLP blocks are still kept unchanged here. Fig.~\ref{p2-c} illustrates the architecture of the X-Alternate layers. This approach is supposed to communicate and extract the information of width and height more sufficiently. Besides, this is equivalent to repeat the \emph{width-cross} and \emph{height-cross} MLP blocks in the basic X-MLP layer twice, except that X-Alternate layer equips the fully connected layers with activation layers and expansion factor. 



\textbf{X-Superior. }In the architecture of X-Superior, we mainly construct more skip connections and  \emph{channel-cross} MLP blocks on the basis of the basic X-MLP layer. Fig.~\ref{p2-d} illustrates the architecture of the X-Superior layers. Specifically, we mix the feature maps of three dimensions straightforwardly by skip connection to enhance the features extraction. In addition, we aggregate the channel information after the feature mix of the three dimensions. This elaborate design enhances both spatial and channel features extraction and attains superior performance on the most benchmark datasets. 



\subsection{Restore the Spatial Weights}\label{s2}
As mentioned above, the \emph{width-cross} and \emph{height-cross} MLP blocks are combined to extract the spatial features. In this section, we will restore the weights for extracting spatial features.

Here, we take the basic X-MLP for example. Since we only take the spatial features into consideration, we omit the channel dimension. Given the input feature map $\textbf{X} \in \$\mathbb{R}^{H\times W}$. For the convenience, we suppose that the size of the output $\textbf{Y}$ is unchanged. 
The weight $\textbf{W}^1 \in \mathbb{R}^{N\times N}$ is learned to interact the information across the width. The weight $\textbf{W}^2 \in \mathbb{R}^{N\times N}$ is learned to interact the information across the height. First we operate the \emph{width-cross} layer on $\textbf{X}$:
\begin{equation}
    \begin{split}
    &\textbf{Y}^{\prime}_{i,j} = \textbf{W}^1_{j,1}\textbf{X}_{i,1}+\textbf{W}^1_{j,2}\textbf{X}_{i,2}+\cdot\cdot\cdot+\textbf{W}^1_{j,W}\textbf{X}_{i,W} \\
    &\textbf{where}\ \ i \in [1,H],\ \ j \in [1,W]
    \label{eq5}
    \end{split}
\end{equation}

Then we conduct the \emph{height-cross} layer to get $\textbf{Y}$:
\begin{equation}
    \begin{split}
    &\textbf{Y}_{i,j} = \textbf{W}^2_{i,1}\textbf{Y}^{\prime}_{1,j}+\textbf{W}^2_{i,2}\textbf{Y}^{\prime}_{2,j}+\cdot\cdot\cdot+\textbf{W}^2_{i,H}\textbf{Y}^{\prime}_{H,j} \\
    &= \textbf{W}^2_{i,1}(\textbf{W}^1_{j,1}\textbf{X}_{1,1}+\textbf{W}^1_{j,2}\textbf{X}_{1,2}+\cdot\cdot\cdot+\textbf{W}^1_{j,W}\textbf{X}_{1,W}) \\
    &+ \textbf{W}^2_{i,2}(\textbf{W}^1_{j,1}\textbf{X}_{2,1}+\textbf{W}^1_{j,2}\textbf{X}_{2,2}+\cdot\cdot\cdot+\textbf{W}^1_{j,W}\textbf{X}_{2,W}) \\
    &+\cdot\cdot\cdot\\
    &+ \textbf{W}^2_{i,H}(\textbf{W}^1_{j,1}\textbf{X}_{H,1}+\textbf{W}^1_{j,2}\textbf{X}_{H,2}+\cdot\cdot\cdot+\textbf{W}^1_{j,W}\textbf{X}_{H,W}) \\
    &\textbf{where}\ \ i \in [1,H],\ \ j \in [1,W]
    \label{eq6}
    \end{split}
\end{equation}
We can rewrite the expression of $\textbf{Y}_{i,j}$ as exhibited below:
\begin{equation}
    \begin{split}
    &\textbf{Y}_{i,j}\!=\!\textbf{W}^2_{i,1}\textbf{W}^1_{j,1}\textbf{X}_{1,1}\!+\!\textbf{W}^2_{i,1}\textbf{W}^1_{j,2}\textbf{X}_{1,2}\!+\!\cdot\cdot\cdot\!+\!\textbf{W}^2_{i,1}\textbf{W}^1_{j,W}\textbf{X}_{1,W} \\
    &\!+\!\textbf{W}^2_{i,2}\textbf{W}^1_{j,1}\textbf{X}_{2,1}\!+\!\textbf{W}^2_{i,2}\textbf{W}^1_{j,2}\textbf{X}_{2,2}\!+\!\cdot\cdot\cdot\!+\!\textbf{W}^2_{i,2}\textbf{W}^1_{j,W}\textbf{X}_{2,W} \\
    &+\!\cdot\cdot\cdot\\
    &+\! \textbf{W}^2_{i,H}\textbf{W}^1_{j,1}\textbf{X}_{H,1}\!+\!\textbf{W}^2_{i,H}\textbf{W}^1_{j,2}\textbf{X}_{H,2}\!+\!\cdot\cdot\cdot\!+\!\textbf{W}^2_{i,H}\textbf{W}^1_{j,W}\textbf{X}_{H,W} \\
    &\textbf{where}\ \ i \in [1,H],\ \ j \in [1,W]
    \label{eq7}
    \end{split}
\end{equation}
The spatial weights restored $\textbf{W}^{\prime} \in \mathbb{R}^{H\times W\times H\times W}$ and the mapping can be expressed as:
\begin{equation}
    \begin{split}
    &\textbf{Y}_{i,j} = \sum_{(a,b)}\textbf{W}^{\prime}_{a,b,i,j}\textbf{X}_{a,b}\\
    &\textbf{W}^{\prime}_{a,b,i,j} = \textbf{W}^2_{i,a}\textbf{W}^1_{j,b}\\
    &\textbf{where}\ \ a,i \in [1,H],\ \ b,j \in [1,W]
    \label{eq8}
    \end{split}
\end{equation}

The restored spatial weights can be regarded as a kind of kernels, which actually aggregate the global neurons of the input and map them to certain pixels in the output. This process bears resemblance to convolution kernels in mathematical expression. However, convolution kernels aggregate the local features, but X-MLP layer extracts the information globally and captures long-range dependencies. In addition, the parameters of convolution kernels are shared across the space but specific across the channels. For X-MLP, the parameters are specific for different positions, but they are shared across the channels. It tends to adopt the spatial-specification to compensate the channel-agnostic ~\cite{li2021involution}. 


\begin{table*}[!ht]
  \caption{Test accuracy of various methods on ten benchmark datasets. The number of parameters and Flops of each method are also listed.}
   \label{t3}
  \centering
   \begin{tabular}{|c|l|c|c|c|c|c|c|c|}
   \hline
   \multicolumn{2}{|c|}{Method}	& Tiny ImageNet & Caltech-256 & Flowers-102	 & Food-101 &	CIFAR-100 &Params (M) & Flops (G)\\
   \hline
   CNN & ConvNet-T  &31.28\% &27.32\%	 &33.76\%	&35.03\%	&51.46\% &35.91	&0.54	\\
   \cline{2-9}
   & CIFAR-Quick  &35.24\%  &32.94\%	&37.50\% &37.41\%	&54.25\%	&36.07	&0.68\\
   \cline{2-9}
   & VGGNet-13  & 37.66\% &33.75\%	&40.02\%	&41.22\%	&57.38\%	&59.60	&1.11\\
   \hline
   MLP &PlainMLP-12   &25.41\%  &27.73\%	&26.81\%	&23.54\%	&42.11\%	&25.76	&1.52	\\
   \cline{2-9}
   &MLPMixer-12   & 33.10\%	&31.21\%&	33.35\%&	34.80\%&	54.89\%&25.76	&1.82	\\
   \cline{2-9}
   &ResMLP-12  &28.28\%&	24.32\%&	20.67\%&	22.87\%&	44.75\%&25.38	&1.77	 \\
   \cline{2-9}
   &PlainMLP-24    &27.84\%  &26.57\%	&27.42\%	&26.86\%	&51.51\%	&51.52	&3.04	\\
   \cline{2-9}
   &MLPMixer-24   & 28.82\%&	31.53\%&	34.75\%&	32.21\%&	54.21\%&51.52	&3.64	\\
   \cline{2-9}
   &ResMLP-24   & 28.31\%&	24.17\%&	21.93\%&	22.54\%&	44.71\%&50.76	&3.54	\\
   \hline
   X-MLP&X-Basic   &35.58\% 	&26.16\% 	&38.09\%	&35.21\%&	52.24\%&13.89	&1.68	\\
   \cline{2-9}
   &X-Exp    & 39.12\%&	31.67\% &	37.78\%&	41.73\%&	50.69\%&14.06	&2.13	\\
   \cline{2-9}
   &X-Alt   & 42.02\%&	\textbf{34.78\%}	&\textbf{40.52\%}&	42.59\%&	\textbf{60.88\%}&14.07	&2.87	 \\
   \cline{2-9}
   &X-Sup   & \textbf{42.32\%} &	32.10\% &	39.18\%&	\textbf{45.14\%}&	52.86\%&17.00	&2.01	\\
   \hline
   \multicolumn{9}{|c|}{ }\\
   \hline
   \multicolumn{2}{|c|}{Method}	& CIFAR-10 &	SVHN & MNIST	 & KMNIST &	Fashion-MNIST &Params (M) & Flops (G)\\
   \hline
   CNN & ConvNet-T  &85.14\% &91.01\%	 &98.78\%	&94.34\%	&89.21\% &35.91	&0.16		\\
   \cline{2-9}
   & CIFAR-Quick  &86.62\%  &91.20\%	 &99.06\%	&94.93\%	&89.86\%	&36.07	&0.20	\\
    \cline{2-9}
   & VGGNet-13  &\textbf{87.57\%}  &91.96\%	&99.16\%	&95.43\%	&90.77\%	&59.60	&0.28	\\
   \hline
   MLP &PlainMLP-12    &81.25\%  &86.11\%	&97.24\%	&89.72\%	&88.41\%	&25.76	&1.52	\\
   \cline{2-9}
   &MLPMixer-12   &83.17\%	&89.27\%&	98.81\%&	92.84\%&	90.79\% &25.76	&1.82	\\
   \cline{2-9}
   &ResMLP-12  &84.27\%&	88.98\%&	98.04\%&	90.51\%&	89.42\%  &25.38	&1.77	 \\
   \cline{2-9}
   &PlainMLP-24    &81.22\%  &88.49\%	&96.92\%	&92.23\%	&88.37\%	&51.52	&3.04	\\
   \cline{2-9}
   &MLPMixer-24   & 82.89\%&	89.47\%&	98.76\%&	93.15\%&	90.65\% &51.52	&3.64	\\
   \cline{2-9}
   &ResMLP-24   & 83.46\%	&89.19\%&	98.07\%&	90.78\%&	89.27\% &50.76	&3.54	\\
   \hline
   X-MLP&X-Basic   & 83.52\%&	90.21\%&	98.55\%&	92.58\%&	90.61\% &13.87	&1.28	\\
   \cline{2-9}
   &X-Exp    & 83.86\%	&91.89\%&	99.12\%&	93.87\%	&91.22\% &13.93	&1.39	\\
   \cline{2-9}
   &X-Alt   & 85.21\%&	\textbf{93.83\%}&	99.09\%&	93.40\%&	91.54\% &13.96	&2.01	 \\
   \cline{2-9}
   &X-Sup    & 85.35\%&	92.14\%&	\textbf{99.21\%}&	\textbf{95.48\%}&	\textbf{91.63\%} &16.98	&1.55	\\
   \hline
   \end{tabular}
\end{table*}

\section{Experiments}

\subsection{Configurations}
This section states the experimental configurations, including the benchmark dataset, compared models, and training details.  

\textbf{Dataset Description.} To make the experimental results more persuasive, we conduct our experiments on ten benchmark datasets. We choose five relatively challenging datasets, including Tiny-ImageNet and Flower-102. We also test the compared models on five classic and widely utilized datasets, like CIFAR-10. All the benchmark datasets are summarized in Table.~\ref{t1}, which exhibits the number of classes, the total amount of the samples in train and test sets, the sizes of the original and input images, training epochs and data augmentation strategy.

\textbf{Compared Methods.}We test X-MLP models equipped with the basic X-MLP layers and other improved X-MLP layers as illustrated above. All the X-MLP models consist of $13$ X-MLP layers. For the convenience, the channel numbers of the feature maps produced in each X-MLP layer is set according to that in VGG-16 ~\cite{simonyan2014very} which composes of 13 convolution layers. The number of the hidden neurons in each \emph{channel-cross} MLP block is set accordingly. The width and the height of the feature maps are reduced by half when the channel number is doubled, which is consistent with the classic design in CNNs. The numbers of the hidden neurons in the \emph{width-cross} and \emph{height-cross} MLP blocks are set accordingly. Besides, for the inputs with different sizes, the lower limit of the final feature map size is set to be $8\times 8$. That is, the feature map will not be reduced when it has been reduced to $8\times 8$. We do not fine tune the hyper-parameters elaborately.

In terms of the currently existing vision MLPs, the PlainMLP~\cite{melas2021you}, MLPMixer ~\cite{tolstikhin2021mlp} and ResMLP ~\cite{touvron2021resmlp} are tested in our experiments. We test these MLP models with 12 or 24 layers. The patch number of these vision MLPs is set to be 16 constantly. The input images are convolutional patch embedded accordingly. Dropout or other random behavior of neuronal activation is not introduced into all MLPs. 

In addition, we compared vision MLPs with some classic CNN models, namely, CIFAR-Quick~\cite{krizhevsky2012imagenet} and VGGNet~\cite{simonyan2014very}. ConvNet-T~\cite{cai2021study} is a basic CNN model which halves the width of VGGNet-9~\cite{simonyan2014very}. The specifications of these compared methods, namely, the number of parameters and Flops, are shown in Table.~\ref{t3}. 

\textbf{Training Details.}To fully explore the effectiveness of different architectures instead of pursuing state-of-the-art performance, we do not utilize any modern strategies like per-training and heavy data augmentation, which have been proven to be beneficial to the results. 
In all the experiments, we train a model on certain datasets from scratch without any extra training data. Specifically, we initialize the learned parameters with Xavier random initialization. 
Cross entropy loss is employed as the loss function. We adopt the SGD algorithm with a batch size of 64,  a momentum of 0.9 and $5\times {10}^{-4}$ weight-decay. Considering currently popular strategies ~\cite{touvron2021resmlp}, employing AdamW optimization with $5\times {10}^{-2}$ weight-decay only makes a slight difference to the final results. 
The numbers of total training epochs on each dataset are exhibited in Table.~\ref{t1}. 
In terms of the learning rate strategy, we set the initial learning rate as $1\times {10}^{-2}$ and decay it by ten times when the training loss is steady. The minimal learning rate is set to be $1\times {10}^{-4}$.
In addition, we still utilized some light data augmentations as a common practice. For CIFAR-10 and CIFAR-100, we pad the image with 4 circles of zero pixels, randomly crop it to the original size, and randomly flip it horizontally with a probability of $0.5$. For Tiny ImageNet, Caltech-256, Flowers-102 and Food-101, we randomly rotate the image within 20 degrees and randomly flip it horizontally with a probability of $0.5$. For the datasets left, we do not utilized any data augmentation. All experimental models are trained on NVIDIA P102 GPU.

\begin{figure*}[!ht]
    \centering
   
    \includegraphics[width=0.98\textwidth]{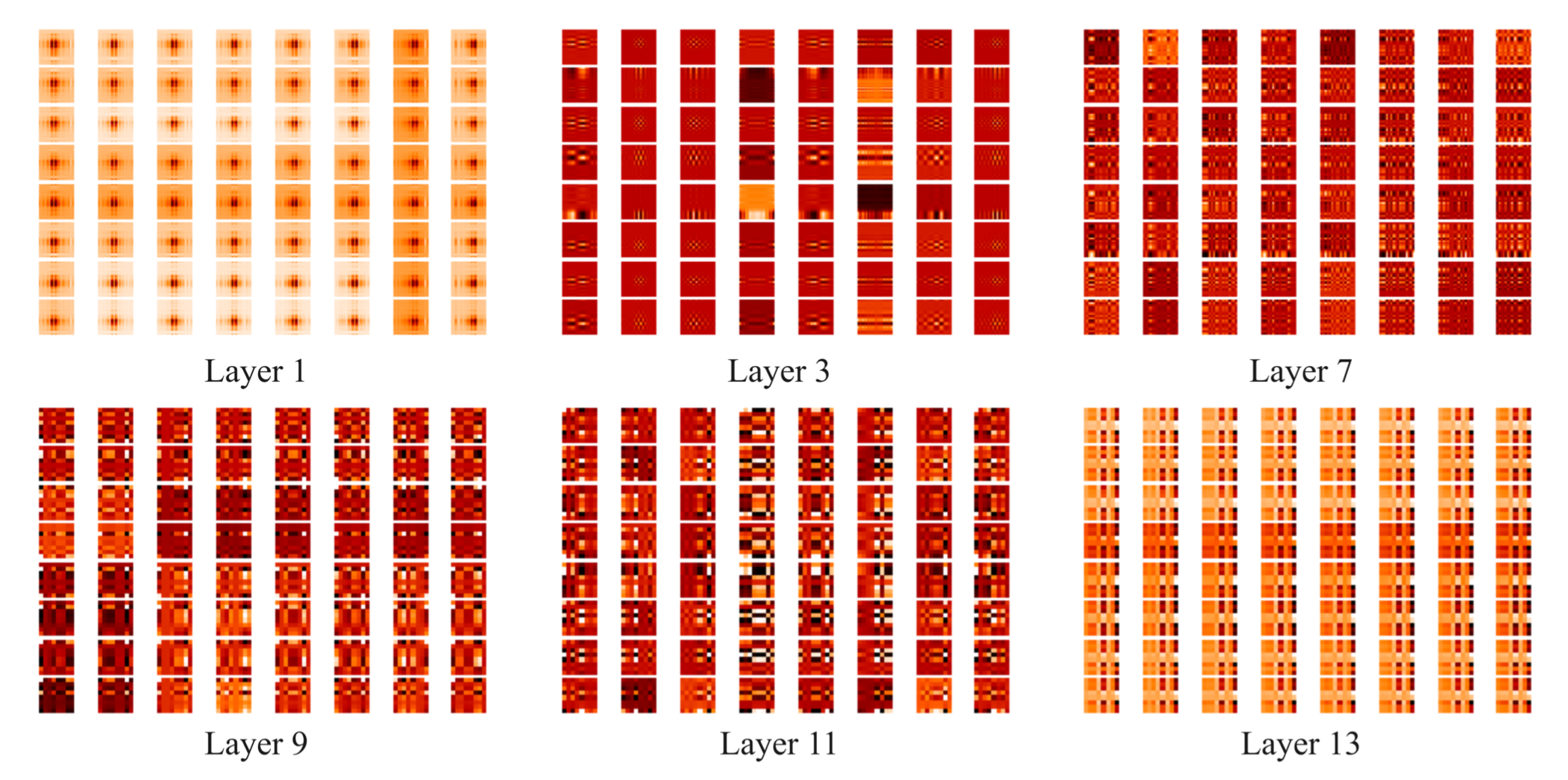}
     
    \caption{Visualization of the restored spatial weights in various layers}
    \label{p3}
\end{figure*}
\subsection{Experimental Results}
We conduct each experiment for five times and calculate the averages. We observe that the test accuracy actually dose not fluctuate significantly. Table.~\ref{t3} shows all the experimental results.

Among all the MLP architecture models, X-Sup and X-Alt are the two models that obtain the best performance.
For X-Sup, it attained the highest test accuracy in six benchmark datasets, namely, Tiny ImageNet, Food-101, CIFAR-10, and three MNISTs. While the baseline MLPMixer-12 is the best model among the other existing vision MLPs, X-Sup surpassed the MLPMixer-12 by $9.22\%$ and $10.34\%$ on Tiny ImageNet and Food-101 respectively, which is significant. Besides, the amount of learned parameters of X-Sup is $1.5$ times fewer than that of MLPMixer-12. 
For X-Alt, it achieved the best performance on the left four datasets. It surpassed the best other existing vison MLPs, MLPMixer-24 and MLPMixer-12, by $5.77\%$ and $5.99\%$ on datasets Flowers-102 and CIFAR-100 with only $54.62\%$ and $27.72\%$ amount of parameters respectively. 
The accuracy gaps between the better X-MLPs and the best existing vision MLPs are $3.25\%$, $1.08\%$, $4.36\%$, $1.40\%$, $2.03\%$, and $0.84\%$ on datasets Caltech-256, CIFAR-10, MNIST, KMNIST, and F-MNIST respectively. These gaps can still be regarded as significant on these datasets. 
For the left two X-MLP models, X-Basic and X-Exp, both of them also achieved relatively good performance. They surpassed other existing vision MLPs on the majority of the datasets with fewer parameters. 


When compared with CNNs, X-MLP still can be compatible and achieve better performance than CNNs on most of the datasets. For example, X-Sup surpasses VGGNet significantly by $5.66\%$ and $3.92\%$ on Tiny ImageNet and Food-101 respectively with only $28.52\$$ amount of parameters. Although it lags behind VGGNet by $1.65\%$ and $0.94\%$ on Caltech-256 and Flowers-102, which is relatively slight.
The experimental results validate the effectiveness of X-MLP architecture.
However, other vision MLPs lag behind significantly on almost all the datasets. This result can illustrate that the modern advances in training and regularization, like pre-training and heavy data augmentation ~\cite{touvron2021resmlp,melas2021you} is vital for these vision MLPs, without which they fail to be compatible with CNNs. 

To conclude, the experimental results prove that X-MLP have better representation ability, which can achieve higher test accuracy with fewer learned parameters, when comparing with other vision MLPs and some classic CNN models, validating the effectiveness of X-MLP architecture persuasively.

\subsection{Visualization}
As stated above, X-MLP layer interacts the information of the width and height dimension respectively. According to Section~\ref{s2}, the weights extracting the spatial features can be restored by combining the width and height weights. As a consequence, the spatial communication between any couples of pixels in the input can be observed clearly. In Fig.~\ref{p1}, we visualize the restored spatial weights sized as $H\times W\times H\times W$ in six chosen X-Basic layers trained on CIFAR-10. For each layer, we select the central $8\times8$ parameters sized as $H\times W$.  
In the first few layers, the patterns obtained are similar with that of convolution. The patterns exhibit a kind of local inductive bias and resemble shifted versions of each other~\cite{touvron2021resmlp}. The following few layers exhibit more complex patterns, containing stripe-like and lattice-like patterns. These patterns also bear resemblance to that in CNNs, as shown in ~\cite{krizhevsky2012imagenet,zeiler2014visualizing}. 
Start from the middle few layers, the patterns become even more complex. It exhibits the intricate global communication among different pixels and the abundant semantic information extracted in the deep layers. The long-range dependencies can be observed clearly.

\section{Conclusion}
In this paper, we raised a novel MLP architecture for visual recognition, termed as X-MLP. X-MLP consists of fully connected layers entirely and has a pyramidal structure. It takes the original images as the inputs and is free from convolutional patch embedding. Moreover, X-MLP introduces a novel way to extract the features which decouples the features extremely. It interacts the information of width, height, and channel dimensions respectively. The global spatial weights can be restored by combining the weights of width and height dimensions. As a consequence, we can observe the communication between any couples of pixels in the input globally and understand the inductive bias by visualization the spatial weights. The experimental results on ten benchmark datasets validated the effectiveness of the X-MLP architecture, which surpassed other existing vision MLPs significantly. 
We hope our work will spark further research on the approach of the feature decoupling, as well as the design of patch embedding-free vision MLPs and ViTs.

\end{document}